%% file: portfolio.tex
\documentclass[]{article}

\usepackage{proceed2e}
\usepackage{times}
\usepackage{natbib}
\usepackage{graphicx}
\usepackage{mwhmacros}

\title{An Entropy Search Portfolio for Bayesian Optimization}

\author{
\begin{tabular}[t]{ccc} 
Bobak Shahriari\thanks{University of British Columbia, Canada} &
Ziyu Wang\thanks{University of Oxford, United Kingdom} &
Matthew W. Hoffman\thanks{University of Cambridge, United Kingdom} \\
\texttt{bshahr@cs.ubc.ca} & \texttt{ziyu.wang@cs.ox.ac.uk} & \texttt{mwh30@cam.ac.uk}\\ 
\end{tabular}
\\
\vspace{0em}
\\
\begin{tabular}{cc} 
\textbf{Alexandre Bouchard-C\^ot\'e}$^\ast$ &
\textbf{Nando de Freitas}$^\dagger$\thanks{Canadian Institute for Advanced Research}\\
\textbf{\texttt{bouchard@stat.ubc.ca}} & \textbf{\texttt{nando@cs.ox.ac.uk}}\\ 
\end{tabular}
}

\def\D{\calD}
\def\X{\calX}

\def\x{\vx}

\def\xrec{\tilde\vx}
\def\xmin{\vx_\star}

\def\P{{\mathbb P}} 
\newcommand{\ud}{\,\mathrm{d}}

\newcounter{xxx}
\renewcommand{\i}[1][i]{^{(#1)}}

\begin{document}

\maketitle

\begin{abstract}
Bayesian optimization is a sample-efficient method for black-box global
optimization. However, the performance of a Bayesian optimization method very
much depends on its exploration strategy, i.e.\ the choice of acquisition
function, and it is not clear \emph{a priori} which choice will result in
superior performance. While portfolio methods provide an effective, principled
way of combining a collection of acquisition functions, they are often based on
measures of past performance which can be misleading. To address this issue, we
introduce the Entropy Search Portfolio (ESP): a novel approach to portfolio
construction which is motivated by information theoretic considerations. We
show that ESP outperforms existing portfolio methods on several real and
synthetic problems, including geostatistical datasets and simulated control
tasks. We not only show that ESP is able to offer performance as good as the
best, \emph{but unknown}, acquisition function, but surprisingly it often gives
better performance.  Finally, over a wide range of conditions we find that ESP
is robust to the inclusion of poor acquisition functions.
\end{abstract}

%===============================================================================
\input{sections/introduction}
\input{sections/portfolio}
\input{sections/algorithm}
\input{sections/experiments}
\input{sections/conclusion}

%===============================================================================
\bibliographystyle{abbrvnat}
\bibliography{bayesopt}

%===============================================================================
\end{document}

%% file: sections/introduction.tex
\section{Introduction}

Bayesian optimization is a popular and successful set of techniques for global
optimization of expensive, black-box functions. These techniques address the
problem of finding the minimizer of a nonlinear function which is generally
non-convex, multi-modal, and whose derivatives are unavailable. Further,
evaluations of the objective function are often only available via
noisy observations. Major applications of these techniques include interactive
user interfaces \citep{Brochu:2010}, robotics \citep{Lizotte:2007,
martinez-cantin:2007}, environmental monitoring \citep{Marchant:2012},
information extraction \citep{Wang:2014aistats}, sensor networks
\citep{Garnett:2010, Srinivas:2010}, adaptive Monte Carlo \citep{Wang:ahmc},
experimental design \citep{Azimi:2012}, and reinforcement learning
\citep{Brochu:2009}. Another broad application area includes the automatic
tuning of machine learning algorithms \citep{Hutter:smac, Bergstra:2011,
Snoek:2012, Swersky:2013, Thornton:2013, Hoffman:2014}.

In general, Bayesian optimization is an iterative optimization procedure which
sequentially queries points and constructs a Bayesian posterior over probable
objective functions. At every step of this procedure the next point to be
evaluated must be selected via some strategy which trades off between exploring
and exploiting our current state of knowledge. Often this strategy is
characterized by means of an acquisition function which explicitly encodes the
value of evaluating any input point. Although from a Bayesian perspective it is
clear that the optimal acquisition function requires multi-step planning, this
is often much too computationally expensive even for short horizons
\citep{osborne:2010}. As a result, current acquisition functions are almost
exclusively myopic measures which approximate the long-term value of a given
evaluation.

% Bobak: Why "almost?" Do you know of any non-myopic acquisition functions? 

One early acquisition function utilized in the literature, Probability of
Improvement (PI), selects the candidate point which has maximum probability of
improving over the value of the best point seen so far (the incumbent). The
point with the highest probability of improvement, however, frequently lies
close to the current incumbent. As a result, the use of PI often leads to
excessively greedy behavior in practice.  Instead, one can quantify the
expected level of improvement, which corresponds to the Expected Improvement
(EI) strategy \citep{Mockus:1978, Jones:1998}. More recently, a wide variety of
more advanced techniques have been proposed which include Bayesian upper
confidence bounds \citep{Srinivas:2010, deFreitas:2012, Wang:2014aistats} and
information-theoretic approaches \citep{Villemonteix:2009, Hennig:2012,
Hernandez:2014}.

% Bobak: Should we refer to our empirical observations in the experiments
% section specifically to make this point stronger?

However, no single acquisition strategy provides better performance over all
problem instances. In fact, it has been empirically observed that the preferred
strategy can change at various stages of the sequential optimization process.
To address this issue, \citep{Hoffman:2011} proposed the use of a portfolio
containing multiple acquisition strategies. A key ingredient of this approach
is a meta-criterion which selects among different strategies and is analogous
to an acquisition function, but at a higher level. Whereas acquisition
functions assign utility to points in the input space, a meta-criterion assigns
utility to candidates suggested by base strategies.

% Bobak: I don't like how this paragraph ends.

Our contribution is a novel meta-criterion which we call the Entropy Search
Portfolio (ESP). While the earlier approach of \citeauthor{Hoffman:2011} relies
on using the past performance of each acquisition function to predict future
performance, our approach instead uses an information-based metric which is
much more suitable when it is unclear which search strategy to use.
Our approach is also closely related to direct, information-based acquisition
functions such as that of \citep{Villemonteix:2009, Hennig:2012,
Hernandez:2014}.

Our proposed method uses, as a subroutine, a technique which produces
approximate sample minimizers of the latent objective function. This approach
relies on an approximation scheme previously used by \cite{Rahimi:2007} and
provides an interesting extension of Thompson sampling, a popular bandit
strategy \citep[see][]{Li:2011} that has so far been
confined to discrete spaces $\calX$. One further contribution in this work, is
to provide the first empirical evaluation of Thompson sampling in continuous
domains on several benchmark global optimization problems.

Fundamentally, the goal of a portfolio strategy is to provide a mechanism which
performs as well as any of its constituent strategies, possibly with some small
loss in efficiency due to the difficulty of identifying the best method.
However, our experimental results will show that ESP not only performs as well
as the base strategies, but it also out-performs both previous portfolio
mechanisms and surprisingly is able to exceed the performance of its base
strategies on many test problems. Finally, since this method does not rely on
the past performances of its constituent strategies, it is more robust to
the inclusion of both poor experts and experts for which the quality of
recommendations degrades over time.

%% file: sections/portfolio.tex
\section{Bayesian Optimization with Portfolios}

As described in the previous section, we are interested in finding the global
minimizer $\xmin=\argmin_{\x\in\X} f(\x)$ of a function $f$ over some bounded
domain, typically $\X\subset\R^d$.  We further assume that $f(\x)$ can only be
evaluated via a series of queries $\x_t$ to a black-box that provides noisy
outputs $y_t$ from some set, typically $\calY \subseteq \R$. For this work we
assume $y_t\sim \Normal(f(\x_t), \sigma^2)$, however, our framework can be
extended to other non-Gaussian likelihoods. In this setting, we describe a
sequential search algorithm that, after $t$ iterations, proposes to evaluate
$f$ at some location $\x_{t+1}$ given by an acquisition strategy $\alpha(\D_t)$
where $\D_t=\{(\x_1,y_1),\dots,(\x_t,y_t)\}$ is the history of previous
observations. Finally, after $T$ iterations the algorithm must make a final
recommendation $\xrec_T$, i.e.\ its best estimate of the optimizer. In this
work we recommend the point which maximizes the posterior mean.

The particular set of strategies we focus on take a Bayesian approach to
modeling the unobserved function $f$ and utilize a posterior distribution over
this function to guide the search process.  In this work, we use a constant-mean
Gaussian process (GP) prior for $f$ \citep{Rasmussen:2006}. This prior is
specified by a positive-definite kernel function $k(\x,\x')$. Given any finite
collection of points $\{\x_1,\dots,\x_t\}$, the values of $f$ at these points
are jointly Gaussian with mean $\mu_0$ and covariance matrix $\vK_t$, where
$[\vK_t]_{ij}=k(\x_i,\x_j)$. For the Gaussian likelihood described above, the
vector of concatenated observations $\vy_t$ is also jointly Gaussian with
mean $\mu_0$. Therefore, at any location $\x$, the latent function $f(\x)$
conditioned on $\D_t$ is Gaussian with marginal mean $\mu_t(\x)$ and variance
$\sigma^2_t(\x)$ given by
\begin{align}
    \mu_t(\x)
    &= \mu_0 + \vk_t(\x)\T(\vK_t+\sigma^2\vI)^{-1}\vy_t\,,
    \label{eq:gpmean}
    \\
    \sigma^2_t(\x)
    &= k(\x,\x) - \vk_t(\x)\T(\vK_t+\sigma^2\vI)^{-1}\vk_t(\x)\,,
    \label{eq:gpvar}
\end{align}
where $\vk_t(\x)$ is a vector of cross-covariance terms between $\x$ and
$\{\x_1,\dots,\x_t\}$.

% Two commonly used kernels that we make use of are the squared exponential and
% Mat\'ern\footnote{Technically the Mat\'ern is a class of kernels; here we use
% a smoothness parameter of $\tfrac52$.} kernels
% \begin{align}
%     k_\textsc{se}(\vx,\vx')
%     &= \nu^2 \exp(-\tfrac12r^2),
%     \\
%     k_\textsc{mat\'ern}(\vx,\vx')
%     &= \nu^2 \exp(-\sqrt5 r) (1+\sqrt{5}r+\tfrac53r^2).
% \end{align}
% where $r = (\vx-\vx')\T\diag(\vell^2)^{-1}(\vx-\vx')$. Both kernels are
% parameterized by $d$ length scale hyperparameters $[\vell]_i$ and an
% amplitude hyperparameter $\nu$. These kernels work well in situations where
% little is known about the space in question, although the Mat\'ern tends to
% make less stringent smoothness assumptions, thus making it a good fit for
% Bayesian optimization.

In this work, we consider a portfolio of base strategies $\mathcal A =
\{\alpha_k\}_{k=1}^K$. Algorithm~\ref{alg:bo} outlines the framework for
Bayesian optimization with portfolios, and the first two panels of
Figure~\ref{fig:pmin} provide a visualization of this process. Each strategy
can be thought of as an expert which recommends that its candidate point
$\x_{kt}=\alpha_k(\D_t)$ be selected at iteration $t$. We will also denote the
$k$th query point as $\x_k$ when the time index is unambiguous. Our task is to
select the most promising candidate according to some meta-criterion, denoted
\textsc{MetaPolicy} in Line~\ref{alg:bo:metapolicy} of the pseudocode. 
The earlier work of \cite{Hoffman:2011} implements a meta-criterion based on
\emph{prediction with expert advice} \citep{Freund:1997}. However, the
performance of this approach relies on the idea that the past performance of
each acquisition strategy is a reasonable predictor of its future performance,
which may not always be the case. Instead, in the next section we will describe
our approach which selects between lower-level strategies based on their
information content.

\begin{algorithm}[t!]
    \caption{Bayesian Optimization with Portfolios}
    \begin{algorithmic}[1]
        \REQUIRE a noisy \textsc{BlackBox} function,\\
                 \hspace{2.6em}initial data $\D_0=\{\}$\\
                 \hspace{2.6em}a portfolio $\calA=\{\alpha_k\}$
                 \vspace{0.5em}
        \FOR{$t=1,\dots,T$}
            \STATE collect candidates $\x_{kt}=\alpha_k(\D_{t-1})$
            \label{alg:bo:portfolio}
            \STATE set $\x_t=\textsc{MetaPolicy}(\{\x_{kt}\}, \D_{t-1})$
            \label{alg:bo:metapolicy}
            \STATE set $y_t=\textsc{BlackBox}(\x_t)$
            \label{alg:bo:observe}
            \STATE augment data $\D_t = \D_{t-1} \cup \{(\x_t, y_t)\}$
            \label{alg:bo:augment}
        \ENDFOR
        \RETURN $\xrec_T=\argmax_{\x\in\X} \mu_T(\x)$
        \label{alg:bo:recommend}
    \end{algorithmic}
    \label{alg:bo}
\end{algorithm}

%% file: sections/algorithm.tex
%==============================================================================
\section{An Entropy Search Portfolio}

Let $\xmin$ denote the \emph{unknown} global minimizer of the latent function
$f$. Given data $\D$, let
\begin{equation}
    \P(\ud \xmin|\D) =
    \P\Big( \argmin_{\x\in\X}f(\x) \in \ud \xmin \Big|\D\Big)
    \label{eq:pmin}
\end{equation}
denote the posterior over minimizer locations with density $p(\xmin|\D)$,
induced by the GP posterior. From this distribution we propose the
meta-criterion
\begin{align}
    u(\x|\D) 
    &=
    \E_{p(y|\D,\x)} \Big[ 
    \H\big[\xmin\big|\D \cup \{(\vx, y)\}\big]
    \Big] 
    \label{eq:utility} \\
    \x_t &= \argmin_{\x_{1t},\dots,\x_{kt}} u(\x|\D)
\end{align}
where $\H$ denotes the entropy functional. In other words the candidate
selected by this criterion is the one that results in the greatest expected
reduction in entropy about the minimizer. If we were to minimize $u(\vx|\D)$ as
a continuous function over the space $\X$ we would arrive at the acquisition
function proposed by \cite{Hennig:2012}. Note, however, that we are instead
restricting this minimization to the set of recommendations made by each
portfolio member. By restricting the set of candidate points we can more
accurately and stably compute the entropy at those points, as we will detail in
the rest of this section.

% In Section~\ref{sec:entropy}, we outline the techniques used to estimate this
% analytically intractable quantity. We will also see that an important step of
% this algorithm relies on sampling from \eqref{eq:pmin}; in
% Section~\ref{sec:sampling}, we show how to approximately sample from this
% distribution, an approach that also acts as an extension of Thompson sampling
% to continuous domains with GP priors. Finally, in
% Section~\ref{sec:algorithm}, we summarize our selection strategy, the
% pseudocode for which is given in Algorithm~\ref{alg:portfolio}. We also
% discuss how to effectively marginalize over GP hyperparameters within this
% strategy.

\begin{figure}[t!]
    \centering
    \includegraphics[width=0.45\textwidth]{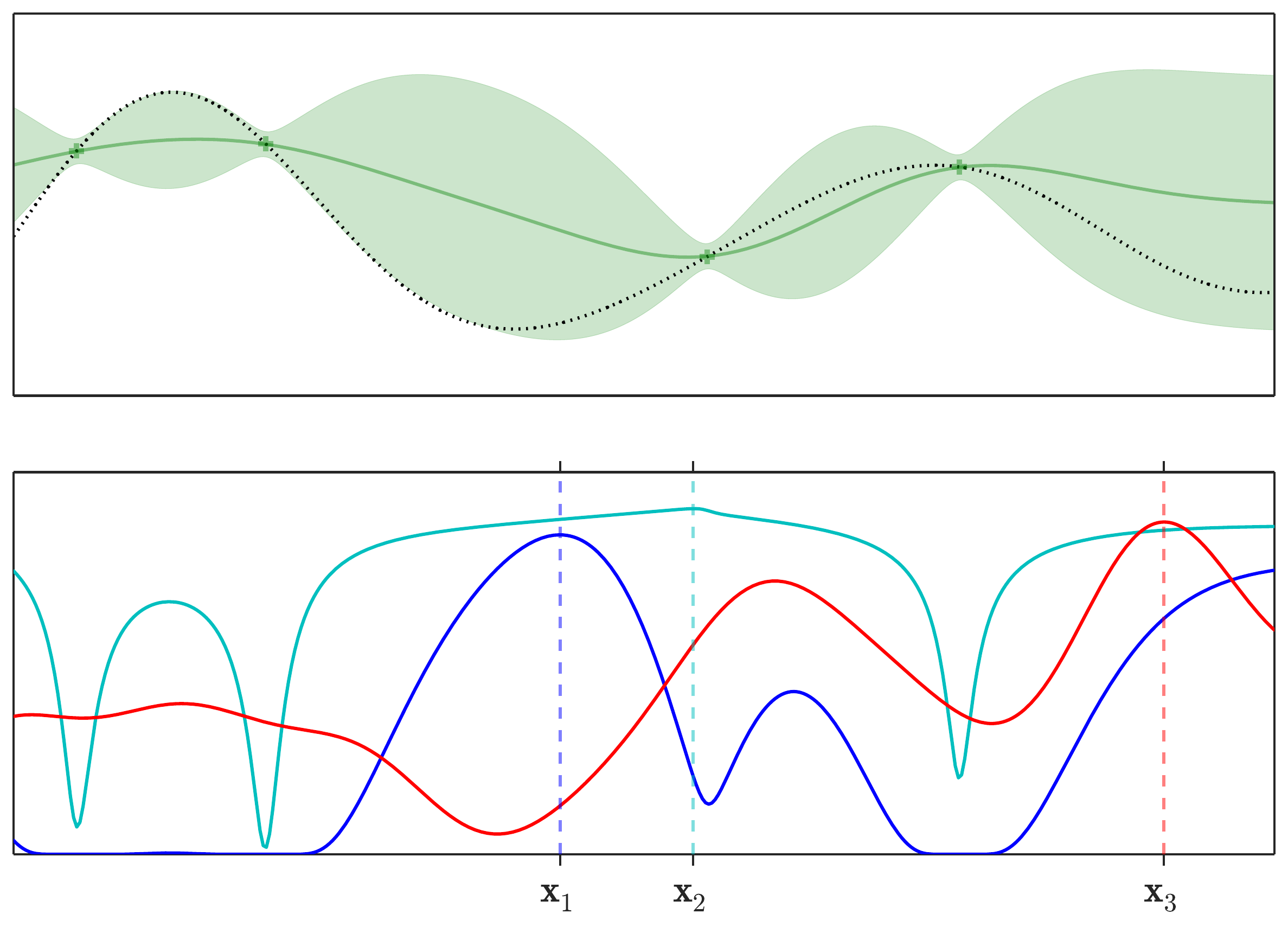}
    \includegraphics[width=0.46\textwidth]{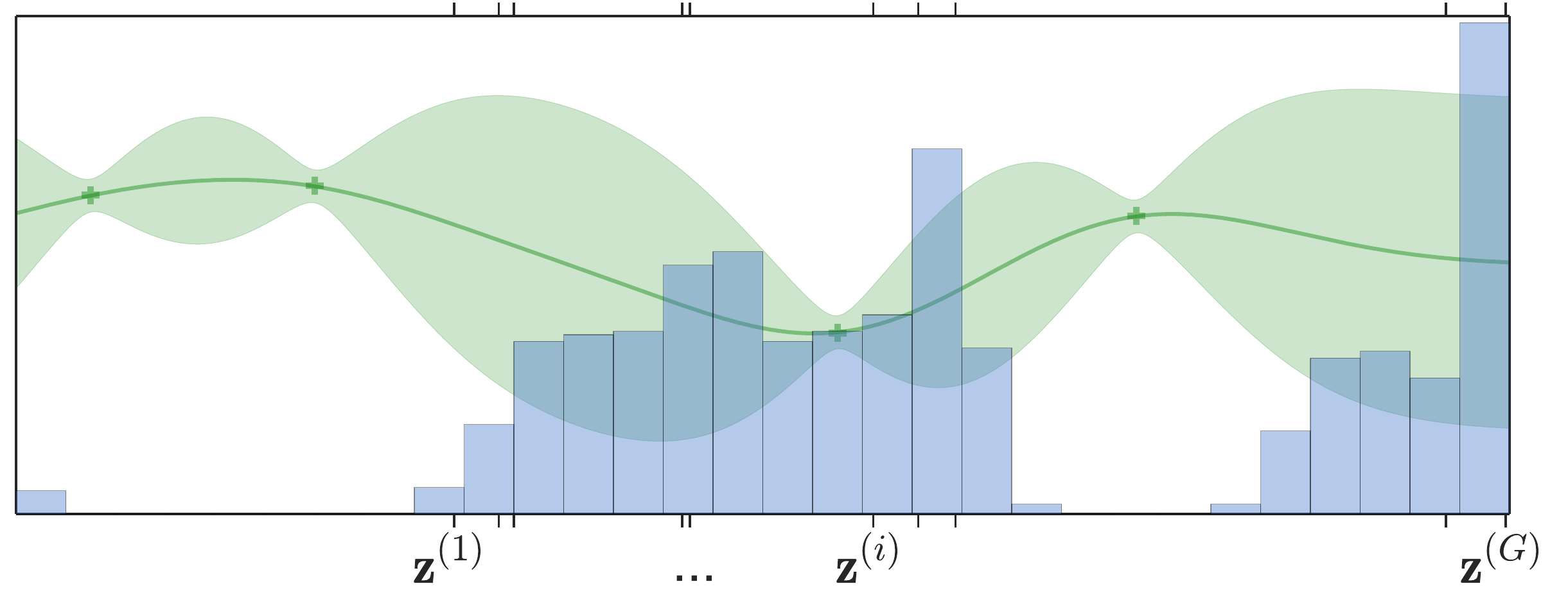}
    \caption{Visualization of Bayesian optimization with acquisition
        function portfolios.
        \textbf{Top panel:} visualization of the marginal GP posterior as well
        as the true objective function (dotted).
        \textbf{Middle panel:} plot of three acquisition functions in this
        portfolio, in this case EI (blue), PI (cyan), and TS (red). The purpose
        of Algorithm~\ref{alg:portfolio} is to select between these
        experts' candidates $\vx_1$, $\vx_2$, and $\vx_3$, respectively.
        \textbf{Bottom panel:} histogram of 1000 approximate samples from
        $p(\xmin|\D)$.}
    \label{fig:pmin}
\end{figure}

\begin{figure}[t!]
    \centering
    \includegraphics[width=0.48\textwidth]{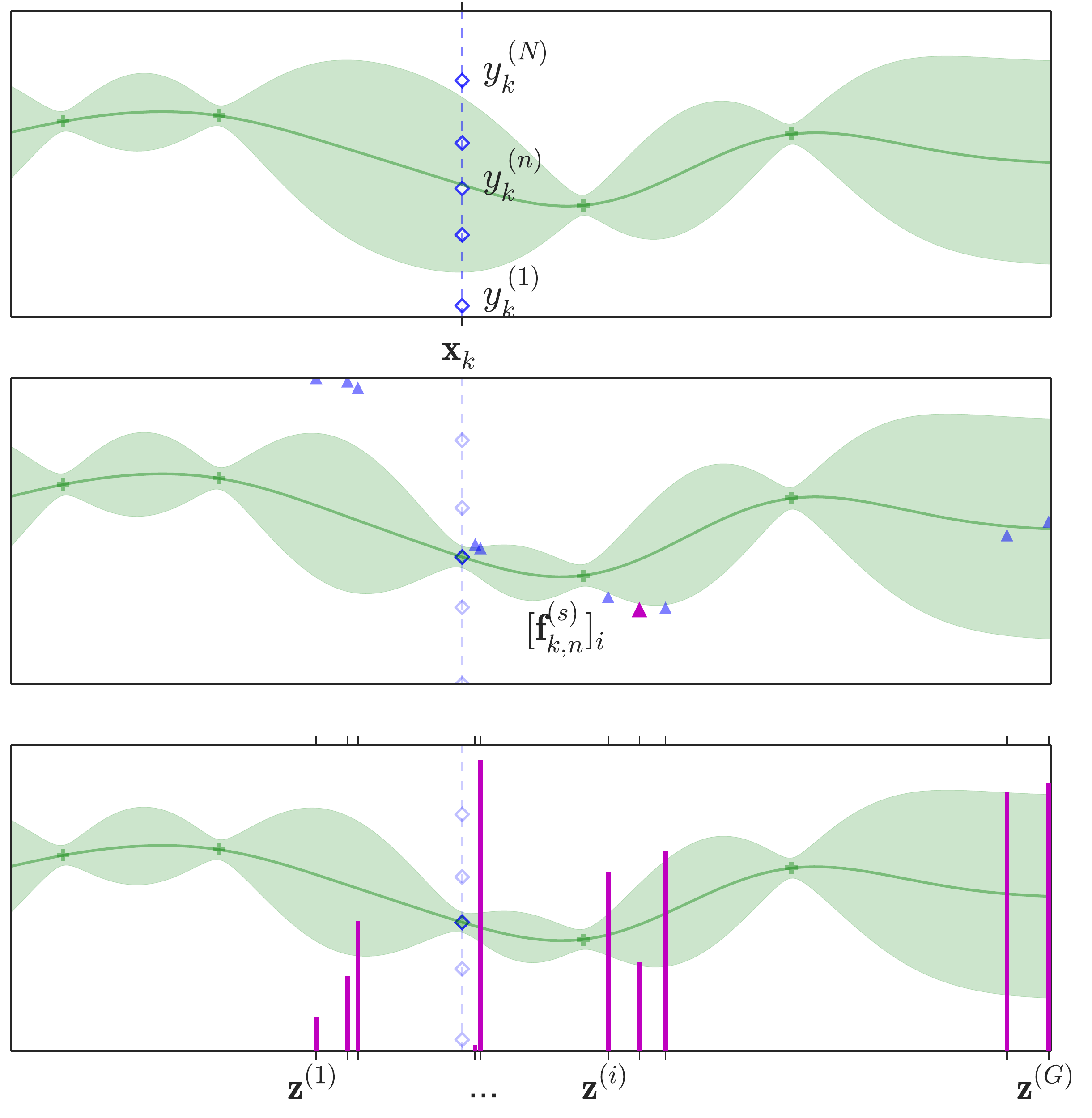}
    \caption{Visualization of the Entropy Search Portfolio.
        \textbf{Top panel:} for each candidate $\vx_k$ (blue dashed line) we
        draw $N$ hallucinations $y_k\i[n]$ (blue diamonds) from the predictive
        distribution.
        \textbf{Middle panel:} for each hallucination we augment the
        GP (green line and shaded area) and sample it $S$ times at the discrete
        points $\vz\i$ to obtain the $\vf_{kn}\i[s]$ (blue triangles) for
        $s=1,\dots,S$. We find the minimizer of each vector $\vf_{kn}\i[s]$
        (magenta triangle).
        \textbf{Bottom panel:} finally, we bin the $S$ minimizers into a
        discrete empirical distribution $\hat p$ depicted here as a magenta
        histogram.}
    \label{fig:metapolicy}
\end{figure}

\subsection{Computing the ESP criterion}

The utility function introduced in \eqref{eq:utility} can be approximated first
by Monte Carlo integration, i.e.\
\begin{align*}
    u(\x_k|\D) 
    \approx \frac1N \sum_{n=1}^N H\big[\xmin|\tilde\D_k\i[n]\big]
    \;\text{for}\;
    y_k\i[n] &\sim p(y|\vx_k,\calD),
\end{align*}
where $\tilde\D_k\i[n] = \calD \cup \{(\vx_k, y_k\i[n]\}$ is the
\emph{hallucinated} data. Note that this step only requires that we sample from
our predictive distribution and in practice one can also use stratified or
quasi-Monte Carlo to reduce the variance of this approximation. Next we must
compute the entropy of the conditional distribution, given by $-\int
p(\xmin|\tilde D)\log p(\xmin|\tilde\D) \ud\x$. However, not only does this
integral not have an analytic solution, but a simple Monte Carlo
approximation is impractical because the quantity $\log p(\xmin|\tilde\D)$
is intractable to compute.

We will instead replace $p(\xmin|\tilde\D)$ with a discrete distribution $\hat
p$ that is restricted to a particular discretization of the domain using a
finite set of \emph{representer points}. These points, denoted
$\{\vz\i\}_{i=1}^G$ will be sampled from some alternative measure as is done
in~\citep{Hennig:2012}. In order to obtain good performance it is desirable for
the distribution of $\{\vz\i\}$ to closely match the target distribution
$p(\xmin|\tilde\D)$. Our proposed algorithm uses approximate samples from the
distribution $p(\xmin|\D)$ which we expect to be closer to the target
distribution than the functions EI or PI that were suggested previously. This
step corresponds to Line~\ref{alg:portfolio:repr} of
Algorithm~\ref{alg:portfolio} after which the $\{\vz\i\}$ are kept fixed for
the remainder of the algorithm. In an effort to maintain the flow of this
section, we defer the technical details of how these approximate samples are
produced to Section~\ref{sec:sampling}.

Combining the Monte Carlo integration with this discretization, the utility for
each recommendation can now be written as
\begin{multline*}
    u(\x_k|\D)
    \approx
    \\
    -\frac1N \sum_{n=1}^N
    \sum_{i=1}^G \hat p(\xmin\!=\!\vz\i|\tilde\D_k\i[n])
                  \log\hat p(\xmin\!=\!\vz\i|\tilde\D_k\i[n]).
\end{multline*}
We are now left with the problem of computing the probability $\hat
p(\xmin=\vz\i|\tilde\D_k\i[n])$. Recall, however, that $\hat p$ is the
distribution over minimizers of a GP-distributed function where the minimizers
are restricted to a finite set. This can easily be sampled exactly. Let the
random variable $[\vf_{kn}]_i=f(\vz\i)$, $i=1,\ldots,G$, be a vector of latent
function values evaluated at the representer points and conditioned on data
$\tilde\D_k\i[n]$. This vector simply has a Gaussian distribution and as a
result we can produce $S$ samples $\vf_{kn}\i[s]\sim p(\cdot|\tilde\D_k\i[n])$
from the resulting GP posterior. The probabilities necessary to compute the
entropy can then be approximated by the relative counts
\begin{equation}
    \hat p_{ikn}=\frac1S
    \sum_{s=1}^S \I\big[i=\argmin_j[\vf_{kn}\i[s]]_j\big]
\end{equation}
such that $\vz\i$ is the minimizer among the sampled functions. Finally, by
combining these ideas we can write our entropy-based meta-criterion as
\begin{equation}
    u(\x_k|\D)
    = -\frac1N \sum_{n=1}^N \sum_{i=1}^G \hat p_{ikn} \log \hat p_{ikn}.
\end{equation}
Pseudocode computing this quantity is given in Algorithm~\ref{alg:portfolio}
and we provide a corresponding visualization in Figure~\ref{fig:metapolicy}.

\begin{algorithm}[t!]
    \caption{Entropy Search Portfolio (ESP)}
    \begin{algorithmic}[1]
    \REQUIRE candidates $\{\x_k\}$, observations $\D$
    \STATE $\vz\i \sim p(\xmin| \D),\quad i=1,\dots,G$
    \label{alg:portfolio:repr}
    \FOR{$k=1:K$}
        \FOR{$n=1:N$}
            \STATE $y_k\i[n] \sim p(y|\x_k, \D)$
            \label{alg:portfolio:simy}
            \STATE $\tilde\D_k\i[n] = \D \cup \{(\x_k, y_k\i[n])\}$
            \STATE $\vf_{kn}\i[s] \sim p(\vf|\tilde\D_k\i[n])$ for $s=1:S$
            \label{alg:portfolio:simf}
            \STATE $\hat p_{ikn} = 
            \frac1S \sum_s\I\big[i=\argmin_j[\vf_{kn}\i[s]]_j\big]$
        \ENDFOR
        \STATE $u_k =
        \frac1N \sum_{n=1}^N\sum_{i=1}^G 
        \hat p_{ikn} \log \hat p_{ikn}$
    \ENDFOR
    \RETURN $\x_{k_\star}$ where $k_\star = \argmax_k u_k$
    \end{algorithmic}
    \label{alg:portfolio}
\end{algorithm}

%==============================================================================
\subsection{Sampling posterior minimizers}
\label{sec:sampling}

The meta-criterion introduced in the previous section relied upon producing
samples from \eqref{eq:pmin}, i.e. the posterior over global minima. While
sampling from \eqref{eq:pmin} is difficult in general, we can gain intuition by
considering the finite domain setting. If the domain $\X$ is restricted to a
finite set of $m$ points, the latent function $f$ takes the form of an
$m$-dimensional vector $\vf$. The probability that the $i$th element of $\vf$
is optimal can then be written as $\int p(\vf|\D) \prod_{j\neq i}
\I[f_i \leq f_j]\,d\vf$. This suggests the following generative process: i)
draw a sample from the posterior distribution $p(\vf|\D)$ and ii) return the
index of the maximum element in the sampled vector. This process is known as
Thompson sampling or probability matching when used as an arm-selection
strategy in multi-armed bandits \citep{Li:2011}. This same approach could be
used for sampling the maximizer over a continuous domain $\X$. At first glance
this would require constructing an infinite-dimensional object representing the
function $f$. To avoid this, one could sequentially construct $f$ while it is
being optimized. However, evaluating such an $f$ would ultimately have cost
$\calO(m^3)$ where $m$ is the number of function evaluations necessary to find
the optimum. Instead, we propose to sample and optimize an analytic
approximation to $f$. We will briefly derive this approximation below.

Given a shift-invariant kernel $k$, a theorem of \citep{Bochner:1959}
asserts the existence of its Fourier dual,
\begin{equation*}
    s(\vw) =
    \frac1{(2\pi)^d} \int e^{i\vw\T\vtau} k(\vtau,\zero) \ud\vtau,
\end{equation*}
i.e.\ the spectral density of $k$.  Letting $p(\vw)=s(\vw)/\alpha$ be the
associated normalized density, we can write the kernel as the expectation
\begin{align}
    k(\x,\x')
    = \alpha \,\E_{p(\vw)}[e^{-i\vw\T(\x-\x')}&]
    \\
    = 2\alpha \,\E_{p(\vw,b)}[\cos(\vw\T\vx+b)&\cos(\vw\T\vx' + b)]\,,
    \label{eq:kernel_approx}
\end{align}
where $b\sim\Uniform[0,2\pi]$. Let $\vphi(\vx)=\sqrt{2\alpha/m} \cos(\vW\x +
\vb)$ denote an $m$-dimensional feature mapping and where $\vW$ and $\vb$
consist of $m$ stacked samples from $p(\vw,b)$. The kernel $k$ can then be
approximated by the inner product of these features, $k(\x,\x')\approx
\vphi(\x)\T\vphi(\x')$. This approach was used by~\citep{Rahimi:2007} as an
approximation method in the context of kernel methods.  The feature mapping
$\vphi(\x)$ allows us to approximate the Gaussian process prior for $f$ with a
linear model $f(\x)=\vphi(\vx)\T\vtheta$ where $\vtheta\sim\Normal(\zero, \vI)$
is a standard Gaussian.  By conditioning on $\D$, the posterior for $\vtheta$
is also multivariate Gaussian, $\vtheta|\D\sim\Normal(\vA^{-1}\vPhi\T \vy,
\sigma^2\vA^{-1})$ where $\vA=\vPhi\T\vPhi+\sigma^2\vI$, $\vy$ is the vector of
the output data, and $\vPhi$ is a matrix of features evaluated on the input
data.

Let $\vphi\i$ and $\vtheta\i$ be a random set of features and the corresponding
posterior weights sampled both according to the generative process given above.
They can then be used to construct the function
$f\i(\x)=\vphi\i(\vx)\T\vtheta\i$, which is an approximate posterior sample of
$f$---albeit one with a finite parameterization. We can then maximize this
function to obtain $\vz\i=\argmin_{\x\in\X} f\i(\vx)$, which is approximately
distributed according to $p(\xmin|\D)$. 

% For the squared-exponential and Mat\'ern kernels we consider here this gives
% rise to weights which take the form
% \begin{align}
%     \vw_\textsc{se}
%     &\sim \Normal(0, \diag(\vell^2)^{-1})
%     \\
%     \vw_\textsc{mat\'ern}
%     &\sim \calT(0, \diag(\vell^2)^{-1}, \tfrac52),
% \end{align}
% i.e.\ these are distributed as a normal and Student's T respectively, which
% are easy to sample from.

Finally, while this approach can be used to produce samples which implement the
entropy meta-criterion of the previous section, it can also be used directly as
an acquisition strategy. Consider the randomized strategy given by
\begin{equation*}
    \alpha_{\textsc{thompson}}(\D) = \xmin\i[0]
    \quad\text{where}\quad
    \xmin\i[0]\sim p(\xmin|\D).
\end{equation*}
Given the above description we can see that for finite $\X$ this is exactly the
Thompson sampling algorithm. However, our derivation extends this approach to
continuously varying functions. Finally we note that this technique was also
used in a similar context by \citep{Hernandez:2014}, albeit not directly as an
acquisition function.

%==============================================================================
\subsection{Additional details}
\label{sec:algorithm}

In Algorithm~\ref{alg:portfolio} we summarize ESP, the meta-criterion developed
in this work. The cost of this approach will be dominated by solving the linear
system on line \ref{alg:portfolio:simf} of Algorithm~\ref{alg:portfolio}. As a
result the complexity will be on the order $\calO(KNSt^3)$, however this is
only a constant-factor slowdown when compared to standard Bayesian optimization
algorithms and can easily be parallelized.

One consideration we have not mentioned is the selection of hyperparameters,
which can have a huge effect on the performance of any Bayesian optimization
algorithm. For example, for kernels with lengthscale parameters chosen to be
very small, then querying a point $ \vx$ will only allow the posterior to learn
the model structure in a very small region around this point. If this parameter
is chosen to be \emph{too small}, this can greatly affect the optimization
process as it will force the optimizer to uniformly explore small balls of
width proportional to the lengthscale. For too large lengthscales, the model
will be too smooth and similar difficulties arise.

Typical approaches to GP regression will often optimize the marginal likelihood
as a means of setting these parameters. However, in the Bayesian optimization
setting this approach is particularly ineffective due to the initial paucity of
data. Even worse, such optimization can also lead to quite severe local maxima
around the initial data points. In this work, we instead consider a fully
Bayesian treatment of the hyperparameters. Let $\vpsi$ denote a vector of
hyperparameters which includes any kernel and likelihood parameters. Let
$p(\vpsi|\D)\propto p(\vpsi) \,p(\D|\vpsi)$ denote the posterior distribution
over these parameters where $p(\vpsi)$ is a hyperprior and $p(\D|\vpsi)$ is the
GP marginal likelihood. For a fully Bayesian treatment of $\vpsi$ we must
marginalize our acquisition strategy with respect to this posterior. The
corresponding integral has no analytic expression and must be approximated
using Monte Carlo.

Consider now drawing $M$ samples $\{\vpsi\i\}$ from the posterior
$p(\vpsi|\D_t)$. Often an acquisition strategy is specified with respect to
some internal \emph{acquisition function} which is optimized at every iteration
in order to select $\x_t$. These functions can then be approximately
marginalized with respect to the hyperparameter samples in order to instead
optimize an integrated acquisition function. This approach was taken in
\citep{Snoek:2012} for the EI and PI acquisition functions. Note that the
randomized Thompson sampling strategy, introduced in
Section~\ref{sec:sampling}, only requires a single hyperparameter sample as we
can see this as a joint sample from the hyperparameters and the function
minimizer.

Our approach has additional complexity in that our candidate selection
criterion depends on the hyperparameter samples as well. This occurs explicitly
in lines (\ref{alg:portfolio:simy}--\ref{alg:portfolio:simf}) of
Algorithm~\ref{alg:portfolio} which sample from the GP posterior. This can be
solved simply by adding an additional loop over the hyperparameter samples. Our
particular use of representer points in line~\ref{alg:portfolio:repr} also
depends on these hyperparameters. We can solve this problem by equally
allocating our representers between the $M$ different hyperparameter samples.

%% file: sections/experiments.tex
\section{Experiments}
\label{sec:experiments}

In this section, we evaluate the proposed method on several problems: two
synthetic test functions commonly used for benchmarking in global optimization,
two real-world datasets from the geostatistics community, and two simulated
control problems. We compare ESP against two other portfolio methods, namely
the Hedge portfolio of \citet{Hoffman:2011} and RP, a random portfolio baseline
which selects among its acquisition functions randomly.

All three portfolios consist of three commonly used base acquisition
functions: EI, PI, and Thompson sampling. For EI we used the implementation
available in the \emph{spearmint}
package\footnote{https://github.com/JasperSnoek/spearmint},
while the latter two were implemented in the same framework. All three methods
were included in the portfolios. Note that we do not compare against GPUCB
(Gaussian Process Upper Confidence Bounds) \cite{Srinivas:2010} as
the bounds do not apply directly when integrating over GP hyperparameters. This
would require an additional derivation which we do not consider here.

The performance of the three constituent acquisition functions are provided in
our figures for three reasons. First, the results show that indeed no single
acquisition function is the top performer across all six objective functions
that we optimized. Second, the purpose of the portfolio was to mitigate against
a poor choice of acquisition function; therefore our target performance should
be the best performing constituent. Surprisingly however, we find that, in two
cases\footnote{Branin and Brenda, and, to a lesser extent, Agromet.}, the
portfolio beat the top acquisition function. Recall that the portfolios do not
know, \emph{a priori}, which of their constituents will perform the best.
Finally, this is also the first evaluation of Thompson sampling in continuous
search spaces, although it was used for a separate purpose in the approximation
method of \citet{Hernandez:2014}.

\subsection{Model details and hyperparameter marginalization}

For this set of experiments we use a Mat\'ern kernel with smoothness parameter
$\tfrac52$. Note that this corresponds to a kernel of the form $k(\vx,\vx') =
\nu^2 \exp(-r) (1+r+\tfrac13r^2)$ where $r = \sqrt5
(\vx-\vx')\T\diag(\vell^2)^{-1}(\vx-\vx')$. Further, for the purposes of
sampling the minimizer locations we need to sample from the spectral density of
this kernel, which can be seen as a Student's T distribution, $\calT(0,
\diag(\vell^2)^{-1}, \tfrac52)$.

As noted in Section~\ref{sec:algorithm}, rather than obtain maximum likelihood
point estimates of the GP hyperparameters (\emph{i.e.,} kernel length-scale and
amplitude, and prior constant mean), we marginalize them using 10 Markov chain
Monte Carlo (MCMC) samples after each observation. Both improvement-based
methods, EI and PI, can simply be evaluated with each sample and averaged.
Meanwhile, in keeping with the Thompson strategy, our Thompson implementation
only uses a single sample, namely the last one of the MCMC chain. As reported
above, our proposed ESP method splits its 500 representer points equally among
the hyperparameter samples. In other words, for each of the 10 MCMC samples we
draw 50 points. In addition, in order to estimate $u(\x_k|\D)$, for each of 5
simulated $y_k\i[n]$ and each GP hyperparameter, ESP draws 1000 samples from
$p(\xmin|\tilde\D_k\i[n])$, and then computes and averages the entropy
estimates.

In the following experiments we evaluate the performance of each algorithm as a
function of the number of function evaluations. When the true optimum is known,
as in the case of the Branin and Hartmann 3 experiments, we measure
performance by the absolute error.
For all other minimization experiments, performance is
given by the minimum function value obtained up to that iteration. Each
experiment was repeated 25 times with different random seeds. For each method,  
we plot the mean performance over the repetitions as well as a shaded area
one standard error away from the mean.

\subsection{Global optimization test functions}

\begin{figure}[t]
    \includegraphics[width=0.49\textwidth]{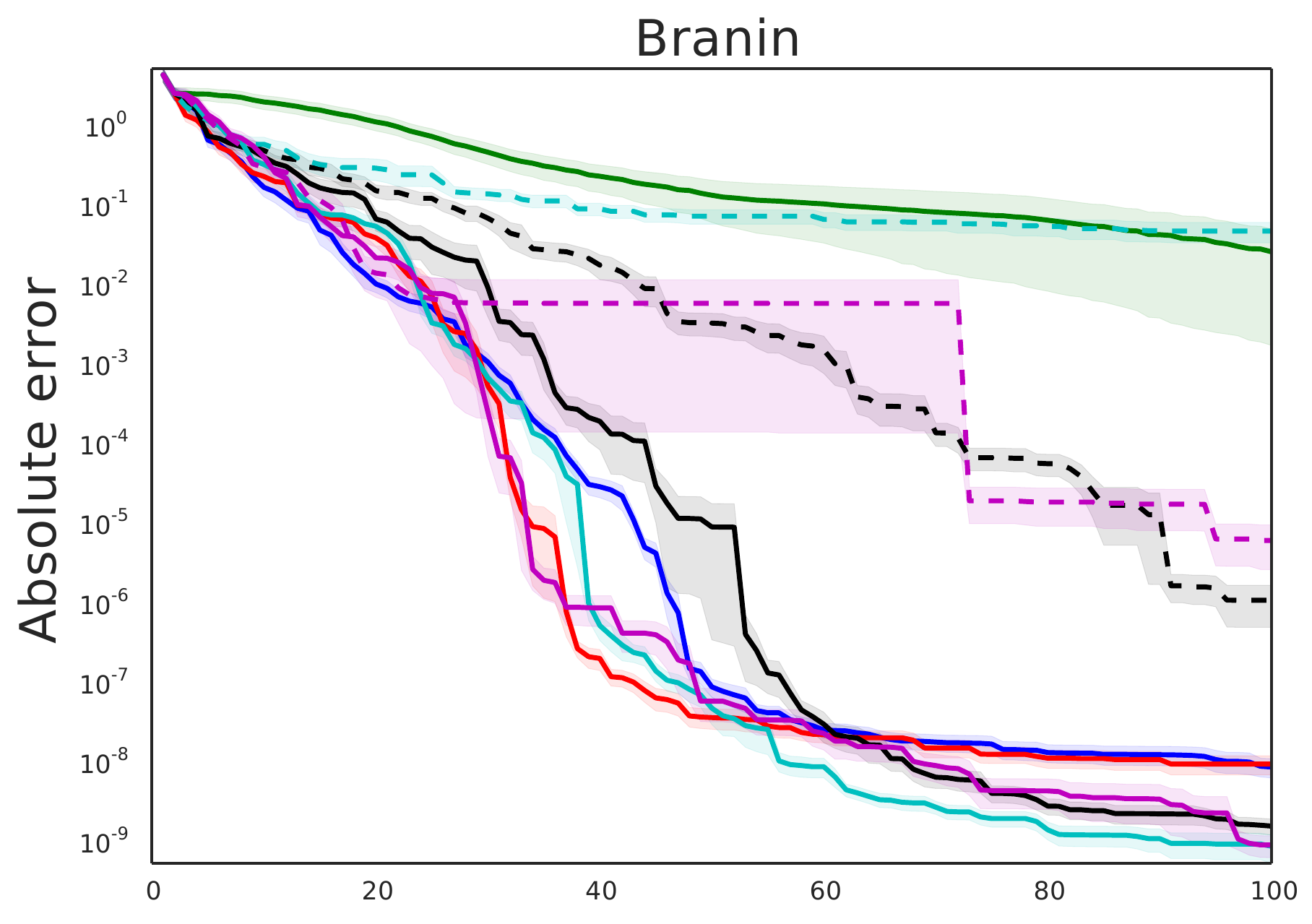}
    \includegraphics[width=0.49\textwidth]{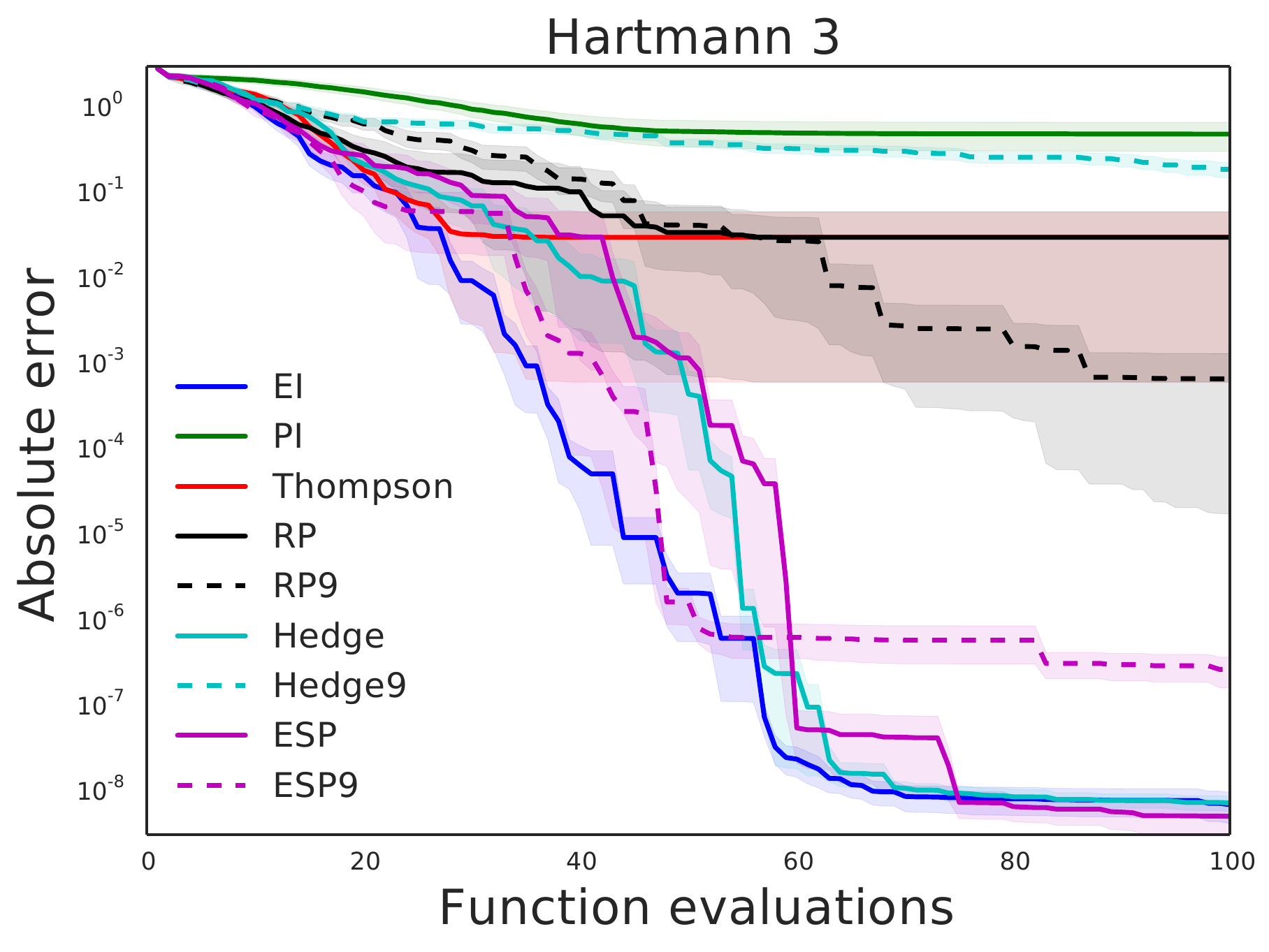}
    \caption{Absolute error of the best observation for the Branin
    and Hartmann 3 synthetic functions. The 9 additional random experts in
    RP9, GPHedge9, and ESP9 affect the RP and GPHedge methods
    much more dramatically than ESP on Hartmann 3. On Branin, a single outlier
	run out of 25 is negatively affecting the averaged results. When this run
	is masked, ESP9 achieves $<10^{-4}$ error by evaluation 40.}
    \label{fig:synthetic}
\end{figure}

We begin with two synthetic functions commonly used for benchmarking
global optimization methods: Branin and Hartmann 3~\citep{Lizotte:2008}.
They are two- and three-dimensional, respectively, and are both continuous,
bounded, and multimodal. The experiments were run up to a final horizon of
$T=100$. Figure~\ref{fig:synthetic} reports the observed performance measured
in absolute error on a logarithmic scale.

It is interesting to note that despite the poor performance of PI in this task,
all portfolios still eventually outperform the base strategies. As expected,
ESP makes the best use of its base experts and outperforms all other methods
in both examples.\footnote{Note that PI performs poorly in this multimodal
example because of PI's greediness. This issue could in principle be addressed
using a \emph{minimum improvement} parameter \cite{Jones:2001,Brochu:2009}.
However we set this parameter to its default value of 0 in our experiments
since our focus in this work is in selecting between acquisition functions
rather than tuning them. However, one could imagine in future work using
parameterized families of strategies as a way of enlarging portfolios.}
In addition, note that each of the two domains have their own champion,
EI being a clear winner on Hartmann 3, and Thompson being the more attractive
option on Branin. This observation motivates the use of portfolios of
acquisition functions.

In these synthetic experiments, we also demonstrate the robustness of ESP with
respect to the inclusion of poor base strategies. We do so by adding 9 random
experts to each portfolio (we denote these ESP9, RP9, etc.). These so-called
random experts select a point uniformly at random in the bounding box $\calX$.
We expect this sort of random search to be comparable to the other base methods
in the initial stage of the optimization and eventually provide too much
exploration and not enough exploitation. Note that a few random experts
in a portfolio could actually be beneficial in providing a constant source of
purely exploratory candidates; precisely how many, however, is an interesting
question we do not discuss in the present work. Nevertheless, for the
dimensionality and difficulty of these examples, we propose 9 additional random
experts as being too many and indeed we observe empirically that they
substantially deteriorate performance for all portfolios.

In particular, we see that, especially on Hartmann 3, ESP is virtually
unaffected until it reaches 6 digits of accuracy. Meanwhile, the progress made
by RP is hindered by the random experts which it selects as frequently as the
legitimate acquisition functions. Significantly worse is the performance of
GPHedge which, due to the initial success of the random experts, favours these
until the horizon is reached.
Note in contrast that ESP does not rely on any expert's past performance, 
which makes it robust to lucky guesses and time-varying expert performances.
GPHedge could possibly be improved by tuning the reward function it uses
internally but this is a difficult and problem dependent task. On Branin, a
single outlier run is deteriorating the performance of ESP9. When masked, the
mean performance achieves $10^{-4}$ by evaluation 40.

\subsection{Geostatistics datasets}

The next set of experiments were carried out on two datasets from the
geostatistics community, referred to here as Brenda and
Agromet~\citep{Clark:2008}.\footnote{Both datasets can
be found at~\texttt{kriging.com/datasets}.} Since these datasets consist in finite 
sets of points, we transformed each of them into a 
function that we can query at arbitrary points via nearest neighbour
interpolation. This produces a jagged piecewise constant function, which is
outside the smoothness class of our surrogate models and hence a relatively
difficult problem.

Brenda is a dataset of 1,856 observations in the depths of a copper mine in
British Columbia, Canada, that was closed in 1990. The search domain is
therefore three-dimensional and the objective function represents the quantity
of underground copper. Agromet is a dataset of 18,188 observations taken in
Natal Highlands, South Africa, where gold grade of the soil is the objective.
The search domain here is two-dimensional.

We note that PI, which has so far been an under-achiever, outperforms EI on
Agromet. This is further motivation for the use of portfolios, because our
experience on the two synthetic experiments would have suggested that PI is a
poor strategy.
On both geological examples, RP is out-performed by GPHedge and ESP.
We observe that ESP performs particularly well on Brenda, and on Agromet
ESP is first, albeit by a very small margin.
Recall that our original intention in using portfolios was to guard against
choosing a poor acquisition function; however, on these examples ESP even
outperforms the best of its constituent strategies (Thompson sampling in these
two examples). This suggests that in addition to improving robustness to poor
acquisition functions, portfolios could be used to boost overall performance.

\begin{figure}[t]
    \includegraphics[width=0.49\textwidth]{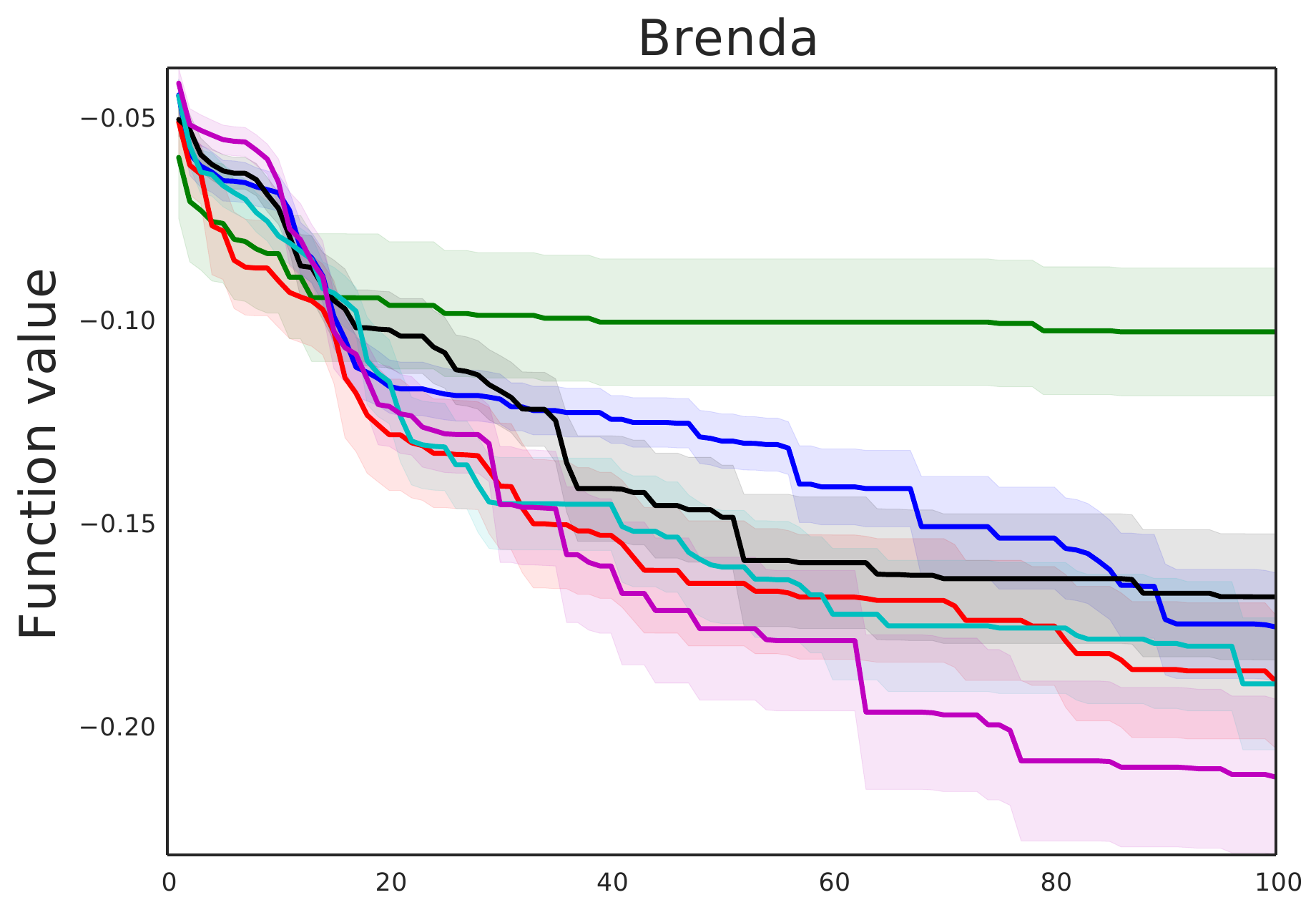}
    \includegraphics[width=0.49\textwidth]{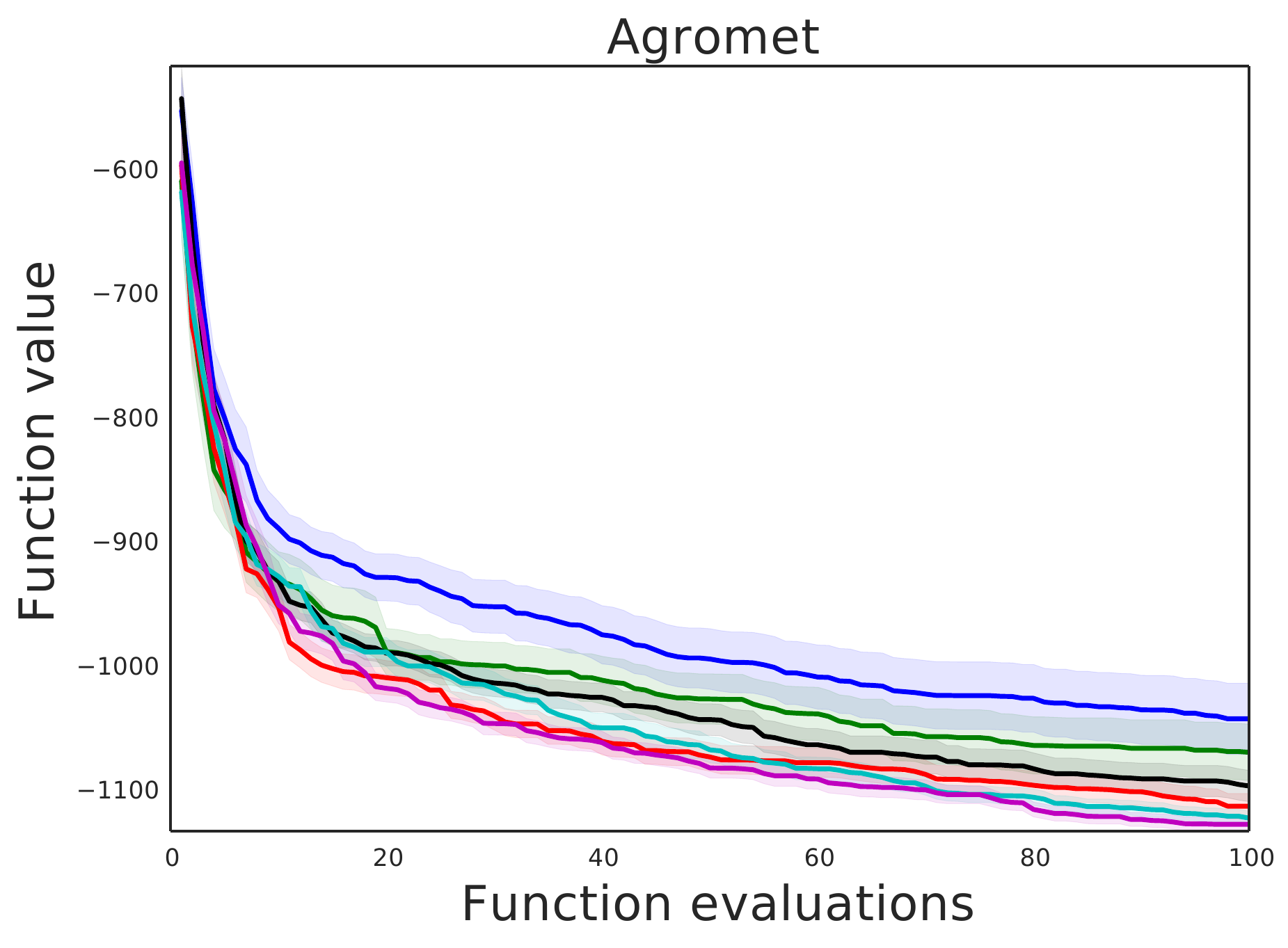}
    \caption{Best observed evaluation on mining datasets Brenda and
    Agromet. ESP outperforms the other portfolio methods while RP
    is dominated by all others.}
    \label{fig:geostats}
\end{figure}

\subsection{Control tasks}

\begin{figure}[t]
    \includegraphics[width=0.49\textwidth]{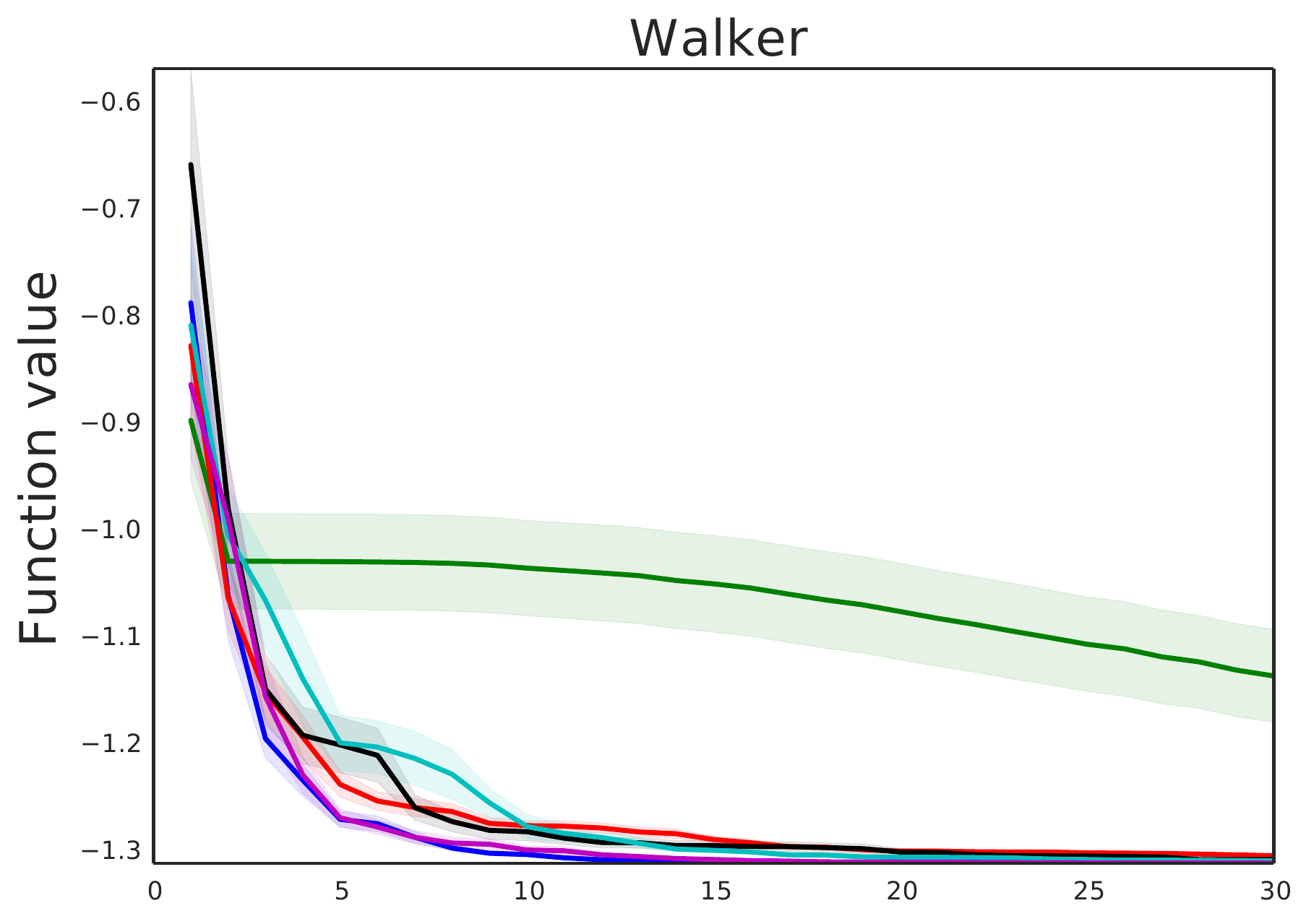}
    \includegraphics[width=0.49\textwidth]{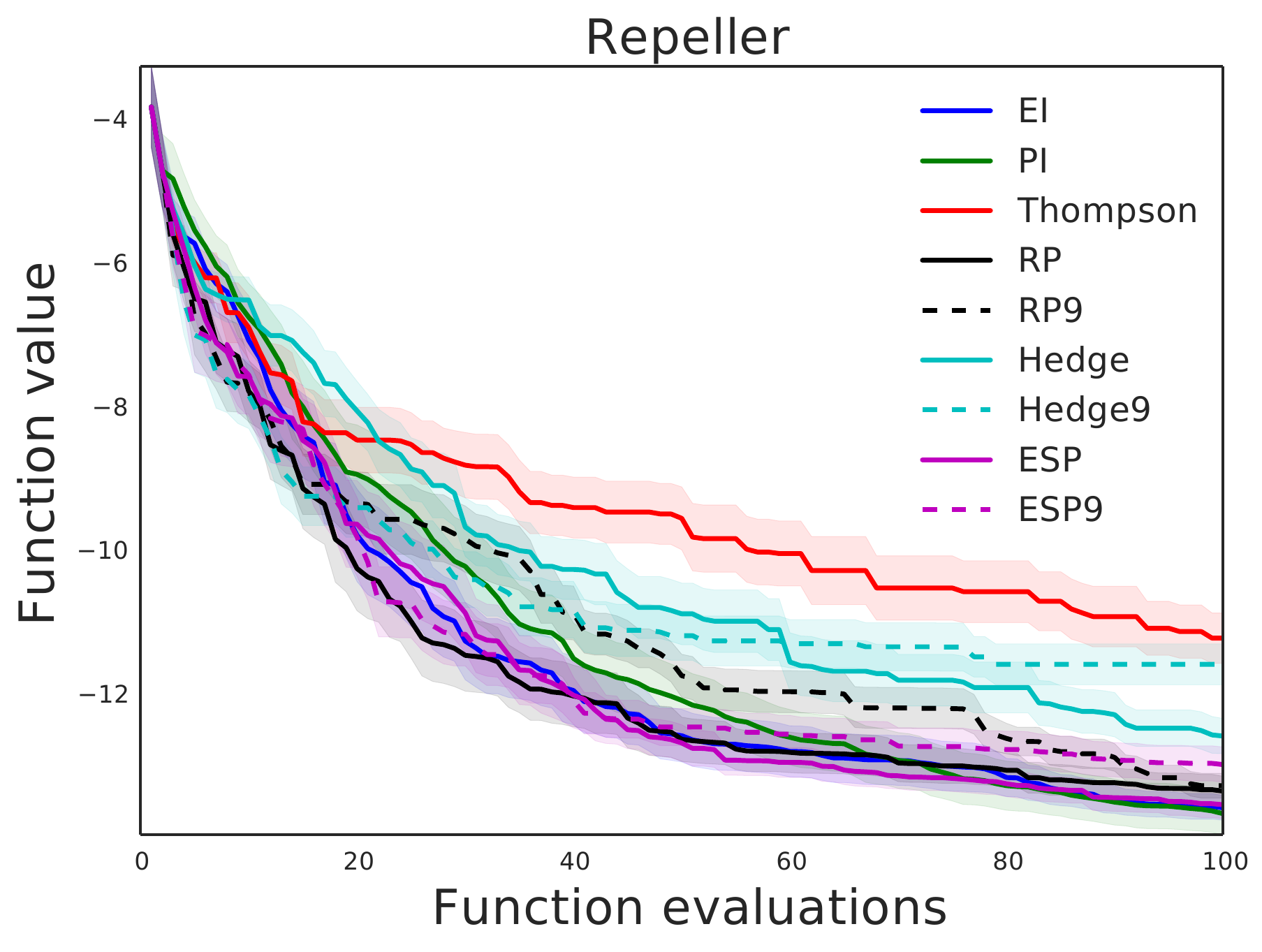}
    \caption{Best observed evaluation on the Walker and Repeller control
    tasks. Here we see ESP outperforming the other methods on Walker. On
    Repeller, ESP is once again competitive with the best methods and
    exhibits much more robustness to poor experts as evidenced by ESP9
    following the performance of ESP so closely.}
    \label{fig:control}
\end{figure}

Our final experiments compare the portfolio methods on two control tasks.
Walker is an eight-dimensional control problem where the inputs are fed into a
simulator which returns the walking speed of a bipedal
robot~\citep{Westervelt:2007}. The code for the simulator is available
online. This example was also used in a Bayesian optimization context by
\citep{Hernandez:2014}.
Repeller is a nine-dimensional control problem considered in previous
work on Bayesian optimisation portfolios~\citep{Hoffman:2011,Hoffman:2009}.
This problem simulates a particle in a two-dimensional world being
dropped from a fixed starting region while accelerated downward by gravity.
The task is to direct the falling particle through circular regions of low
loss by placing 3 repellers with 3 degrees of freedom each, namely their
position in the plane and repulsion strength. Figure~\ref{fig:control}
reports our results on both control tasks.

On these tasks, GPHedge performed poorly, while RP performs well.
Meanwhile, ESP is the clear top performing portfolio method on Walker
and competitive, if not tied, with RP on Repeller.
We have also added the RP9, GPHedge9, and ESP9 methods in this
experiment to once again demonstrate the robustness of ESP relative to its
competitors. Indeed, both RP9 and GPHedge9 exhibit noticeably poorer
performance than RP and GPHedge, while in contrast ESP9 is not significantly
affected.

%% file: sections/conclusion.tex
\section{Conclusion}

In this work we revisited the use of portfolios for Bayesian optimization. We
introduced a novel, information-theoretic meta-criterion ESP which can indeed
provide performance matching or even exceeding that of its component experts.
This is particularly important since we show in our experiments that the
best acquisition function varies between problem instances and is
\emph{a priori} unknown. We have also
shown that ESP has robust behavior across functions of different dimensionality
even when the members of its portfolio do not exhibit this
behavior. Furthermore, ESP is more robust to poorly performing experts than
other portfolio mechanisms.

% We also showed that, surprisingly, a random portfolio can exhibit
% reasonably good performance, albeit not as good as ESP. Such a strategy is also
% more susceptible to the inclusion of poorly performing experts than ESP,
% however not as susceptible as GPHedge due to the latter algorithm's reliance on
% past performance as a predictor of future performance.

Finally, we provided a mechanism for sampling representer points as a way of
approximating $p(\xmin|\D)$ which is more principled than previous approaches
which relied on slice sampling from surrogate measures. We additionally showed
that this sampling mechanism can itself be used as an acquisition strategy,
thereby extending the popular Thompson sampling approach to continuous spaces;
we include the first empirical evaluation of this method within our results.